\documentclass{article} % For LaTeX2e
\usepackage{iclr_modified,times}

\usepackage{latexsym}
\usepackage{amsmath}
\usepackage{multirow}
\usepackage{hyperref}
\usepackage{url}
\usepackage{amssymb}
\usepackage{pgfplots}
\usepackage{soul}
\usepackage{booktabs}
\usepackage[colorinlistoftodos,prependcaption,textsize=tiny]{todonotes}
\usepackage{tikz,tikz-qtree}
\usepackage{multirow}
\usepackage{xspace} 
\usepackage{subcaption} 

\usepackage{wrapfig}


\newcommand{\rectangle}{\fboxsep0pt\fbox{\rule{1em}{0pt}\rule{0pt}{1ex}}}
\newcommand{\hlr}{\overrightarrow{h}}
\newcommand{\hrl}{\overleftarrow{h}}

\newcommand{\plr}{\overrightarrow{p}}
\newcommand{\prl}{\overleftarrow{p}}
\newcommand{\eat}[1]{\ignorespaces}

\definecolor{g-red}{HTML}{DB4437}
\definecolor{g-blue}{HTML}{4285F4}
\definecolor{g-green}{HTML}{0F9D58}
\definecolor{g-yellow}{HTML}{F4B400}
\definecolor{g-orange}{HTML}{FF9800}
\definecolor{g-grey}{HTML}{9E9E9E}

\newcommand\bilstm{{\sc bi}-HLSTM\xspace}
\newcommand\lrlstm{{\sc l$+$r}-HLSTM\xspace}

\newcommand\llm{{\sc l-}LM\xspace}

\newcommand\lm{{\sc l$+$r-}LM\xspace}
\newcommand\loclm{$\text{\lm}^{\text{local}}$\xspace}

\newcommand\glm{$\text{\lm}^{\text{global}}$\xspace}

\newcommand\mlm{{\sc mask-}LM\xspace}
\newcommand\gmlm{$\text{\mlm}^{\text{global}}$\xspace}
\newcommand\elmop{{\sc ELMo}$_{\text{pool}}$\xspace}

\definecolor{antiquewhite}{rgb}{0.98, 0.92, 0.84}

\DeclareMathOperator*{\E}{\mathbb{E}}

\newcommand{\ignore}[1]{}
\newcommand{\ra}[1]{\renewcommand{\arraystretch}{#1}}

\newcommand{\newcite}{\citet}
\renewcommand{\cite}{\citep}

\title{Language Model Pre-training for Hierarchical Document Representations}

\author{Ming-Wei Chang, Kristina Toutanova, Kenton Lee, Jacob Devlin\\
Google AI Language\\
\texttt{\{mingweichang, kristout, kentonl, jacobdevlin\}@google.com} \\
}

\date{}

\iclrfinalcopy 
\begin{document}
\maketitle
\begin{abstract}
Hierarchical neural architectures are often used to capture long-distance dependencies and have been applied to many document-level tasks such as summarization, document segmentation, and sentiment analysis. However, effective usage of such a large context can be difficult to learn, especially in the case where there is limited labeled data available.
Building on the recent success of language model pretraining methods for learning flat representations of text, we propose algorithms for pre-training hierarchical document representations from unlabeled data. Unlike prior work, which has focused on pre-training contextual token representations or context-independent {sentence/paragraph} representations, our hierarchical document representations include fixed-length sentence/paragraph representations which integrate contextual information from the entire documents. 
Experiments on document segmentation, document-level question answering, and extractive document summarization
demonstrate the effectiveness
of the proposed pre-training algorithms.

\end{abstract}

\section{Introduction}

While many natural language processing (NLP) tasks have been posed as isolated prediction problems with limited context (often a single sentence or paragraph), this does not reflect how humans understand natural language. When reading text, humans are sensitive to much more context, such as the rest of the document or even other relevant documents. In this paper, we focus on tasks that require document-level understanding.
We build upon two existing separate lines of work that are useful for these tasks: (1) document-level models with hierarchical architectures, which include sentence representations contextualized with respect to entire documents~\cite{ruder2016hierarchical,cheng-lapata:2016:P16-1,koshorek-etal:2018,Yang2016HierarchicalAN}, and (2) contextual representation learning with language-model pretraining~\cite{peters-etal:2018:_deep, radford-etal:2018}.

In this work, we show for the first time that these two ideas can be successfully combined.
We find that pretraining hierarchical document representations raises
several unique challenges compared to existing representation learning. First, it is not clear how to pretrain the higher level representations.
Using existing techniques, it is possible to initialize the lower level of the hierarchical representation, to represent words in the context of individual sentences e.g. using ELMo~\cite{peters-etal:2018:_deep}, or, in addition, initialize sentence vectors that are not dependent on the context from the full document using Skip-Thought vectors \cite{kiros-etal:2015:_skip}, but it is not possible to initialize the full hierarchical neural structure.
Second, bidirectional representations are standard in state-of-the-art NLP tasks \cite{wu-etal:2016:_googl, weissenborn-wiese-seiffe:2017:CoNLL}, and may be particularly important for document-level representations due to the long-distance dependencies. However, current pretraining algorithms often use left-to-right
and right-to-left language models to pretrain the representations. This restricts the expressiveness of the hierarchical representations, as left and right contextual information do not fuse together when forming higher-level hierarchical representations.

In this paper, we address these challenges by proposing two novel approaches to pre-train document-level hierarchical representations. Both methods pre-train representations 
from unlabeled documents containing thousands
of tokens. Our first approach generalizes the method of \citet{peters-etal:2018:_deep} to pre-train hierarchical left-to-right and right-to-left document representations.
To allow the hierarchical representations to learn to fuse left and right contextual information from abundant unlabeled text, 
our second approach extends the {\em masked language model} technique~\cite{devlin2018bert} to efficiently pre-train bidirectional hierarchical
document-level representations.

We evaluate the impact of the novel aspects of our pre-trained representations on three tasks: document segmentation, answer passage retrieval for
document-level question answering, and extractive text summarization. We first pretrain document-level representations on unlabeled
documents and then fine-tune with task-specific labeled data and a light-weight task  network. Experiments show that pre-training hierarchical document representations
brings significant benefits on all tasks, and that pretraining the higher level of the document representation
results in larger improvements than pre-training only the locally contextualized lower word level in most tasks. 
On the CNN/Daily Mail summarization task, we obtain strong results without using reinforcement learning methods employed by the previous state-of-the-art models.
On the TriviaQA answer passage retrieval task, our model improves over prior work \cite{clark-gardner:2018:_simpl} by 6\% absolute.

\section{Background}
In this section, we review pretraining of contextual token representations using language models and discuss how the language model probability decomposition imposes constraints on the architectures.

\paragraph{Language Model Pretraining}

Given a sequence of tokens $(x_1, \ldots, x_n)$, a representation module $V_\theta $ encodes each word into a vector: $V_\theta(x_1, \ldots, x_n) = (v_1, \ldots, v_n)$,
where each word-level contextual representation $v_i$ could potentially depend on the whole sequence. Common choices for the representation modules are LSTMs,
CNNs, and self-attention architectures~\cite{hochreiter-schmidhuber:1997:_long,LeCun:1998:CNI:303568.303704,vaswani-etal:2017:_atten}. 
A representation module is parameterized by an underlying neural architecture with parameters $\theta$.

The typical left-to-right language model aims to maximize the probability of observed sequences: 
\begin{equation}
\prod_{t=1}^n P(x_t|x_{1}, \ldots x_{t-1}; \theta),
\label{eq:lr_lm_obj}
\end{equation}
The pretrained parameters $\theta$ can then be used to initialize the model for a downstream task.

\paragraph{Uni-directionality constraint}
The formulation of the language model implicitly imposes a directionality
constraint on the $V_\theta$, which we term {\em uni-directionality constraint}.
In Equation~\ref{eq:lr_lm_obj}, the word $x_t$ can only depend on the (representations of) the previous words. Therefore,
the contextual representations $v_{1}, \ldots, v_{t-1}$ can not depend
on the representations of $x_{t}, \ldots, x_{n}$, resulting in the uni-directionality constraint on the representations.

One common type of representation modules that satisfy the uni-directionality constraint 
are uni-directional recurrent neural networks (RNNs). Given a sequence of words, a
left-to-right LSTM model generates a sequence
of representations $V_\theta(x_1, \ldots, x_n) = (\hlr_1,\ldots, \hlr_n$)
that satisfies the uni-directionality constraints, where $\hlr_t$ only
depends on $\hlr_1, \hlr_2, \ldots \hlr_{t-1}$. 
Modules that satisfy the uni-directionality constraint can also be derived from self-attention, convolution, and feed-forward architectures (e.g. \cite{radford-etal:2018}).

As mentioned in the introduction, for many downstream tasks, bidirectional representations bring large improvements over uni-directional ones.  Therefore, \newcite{peters-etal:2018:_deep},
pre-trained two encoders using two language models: a left-to-right and a right-to-left model, using a left-to-right and a right-to-left LSTM encoder, respectively. The token representations are formed by the concatenation of outputs from the two uni-directional encoders.
We term this style of language model pre-training \lm.
More specifically, the representation of the $t$-th word becomes
$v_t = \begin{bmatrix} \hlr_{t} &\hrl_{t} \end{bmatrix}$,
where  $\hrl_t$ is generated from a right-to-left LSTM. Tasks using these input representations must learn to merge the left and right contextual information from scratch using only downstream task labeled data.

The general strategy we adopt in this paper is to pretrain the hierarchical representations on unlabeled data and then to fine-tune the representations with task-specific labeled data. Therefore, the uni-directionality constraint also impacts the task-specific networks.

\begin{figure*}[t]
\hspace{-2cm}
\centering
\begin{subfigure}[b]{0.48\textwidth}
\includegraphics[width=1.2\textwidth]{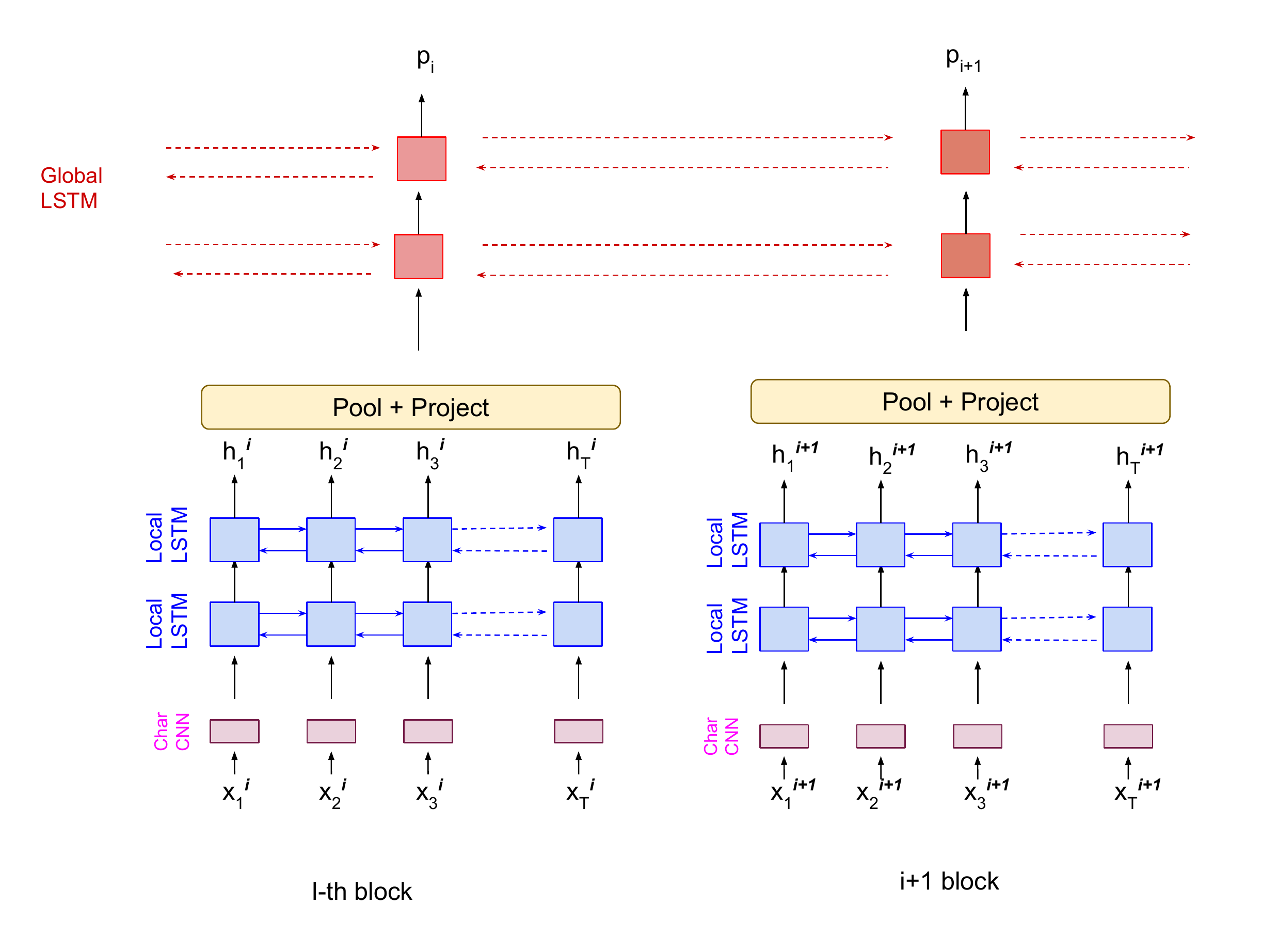}
\caption{\bilstm}
\label{fig:bihlstm}
\end{subfigure}
\hspace{.7cm}
\begin{subfigure}[b]{0.48\textwidth}
\includegraphics[width=1.2\textwidth]{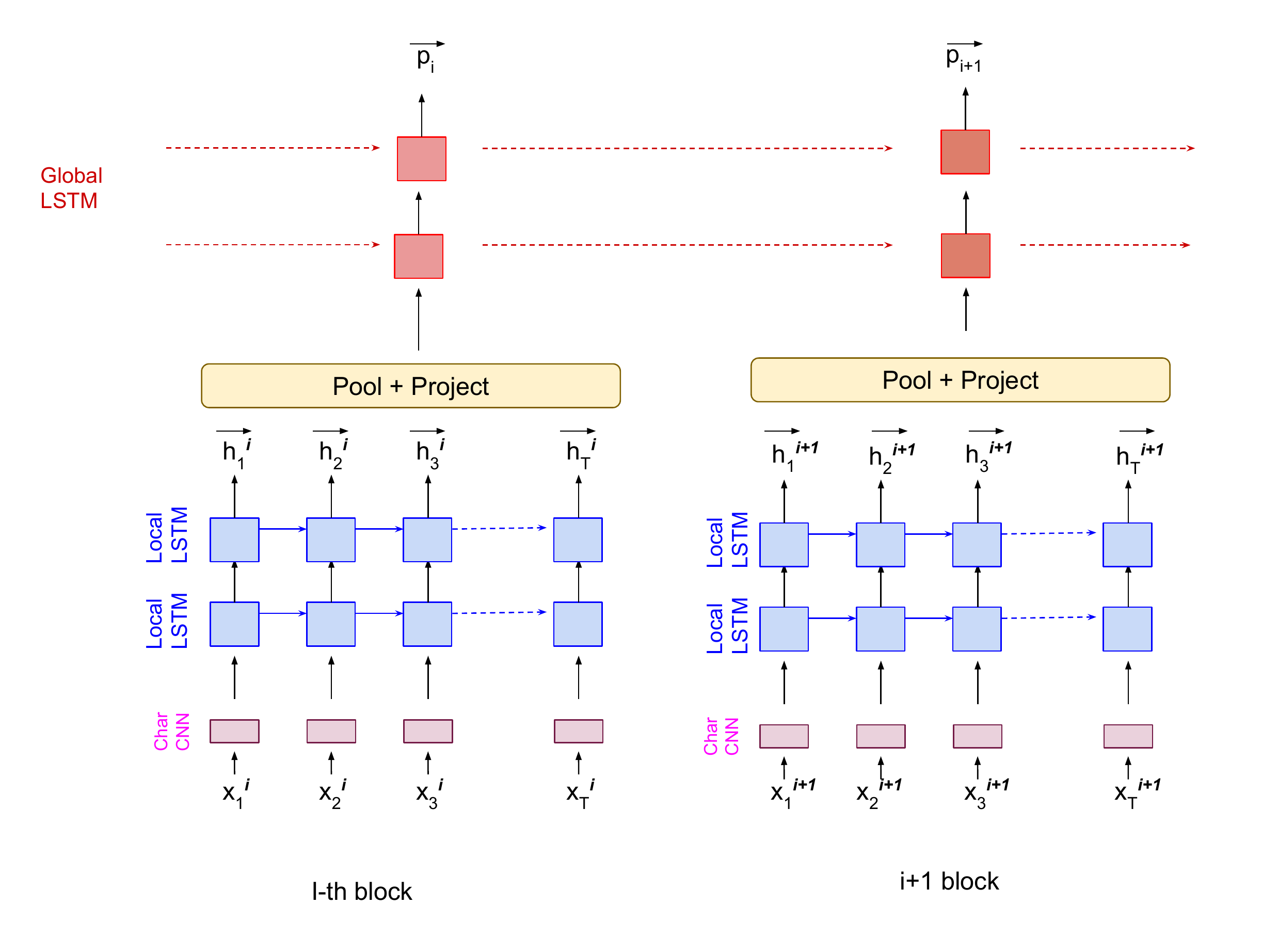}
\caption{left-to-right HLSTM}
\label{fig:left-to-right-hlstm}
\end{subfigure}
\caption{Document-level representations used in this paper. {\bf (a)} Bi-HLSTM:
the bidirectional hierarchical representation, which does not satisfy the uni-directionality constraint. {\bf (b)} left-to-right HLSTM (\textsc{L-to-R-HLSTM}):
a uni-directional version of \bilstm.  }
\end{figure*}

\section{Hierarchical Document-Level Representations}
\label{sec:doc_encoder}

As our goal is to learn fixed-length contextualized representations of text segments to be used in document-level tasks, we choose hierarchical neural architectures that build representations of tokens and text segments in context. Here we describe the specific architectures studied in this paper. The novel contribution of our work -- pre-training such representations from unlabeled text, will be detailed in Section \ref{sec:pre-training}.

We choose LSTM-based architectures with a standard hierarchical structure that has been useful for capturing long-term context in document-level tasks~\cite{serban-etal:2016, koshorek-etal:2018}. We experiment with two related document-level representations. The first one, called \bilstm, is a standard
hierarchical bidirectional LSTM encoder. \bilstm does not satisfy the uni-directionality constraint
and cannot be pre-trained using a left-to-right language model likelihood objective function. 
Our second document-level representation is \lrlstm, which consists
of two concatenated uni-directional versions of
\bilstm, and can be seen as a hierarchical document-level extension of ELMo~\cite{peters-etal:2018:_deep}. Both document-level representations view documents as sequences of text blocks, each consisting of a sequence of tokens. The text blocks can be sentences, paragraphs, or sections, depending on the task. Here we experiment with sentences and paragraphs as text blocks.

The \bilstm architecture fuses left and right contextual information more effectively and directly corresponds to state-of-the-art architectures used for multiple NLP tasks.

\paragraph{\bilstm}
\bilstm has a two-level hierarchy.
Every block is first independently processed  by a bidirectional LSTM (Bi-LSTM) to encode the local context. The contextual representations of the words in each block are then pooled and projected to a single vector to represent the block.
The vector representations of the sequence of blocks in the document are then processed by a global Bi-LSTM, resulting in block representations contextualized by the entire document.
The overall design is shown in Figure~\ref{fig:bihlstm}, and the encoding procedure is formalized next.

A document $D$ consists of $K$ blocks $D = \{U_1, U_2, \ldots U_K\}$, and
each block consists of a sequence of tokens where
$U_i = \{x^i_1, x^i_2, \ldots x^i_T\}$.\footnote{In our experiments, the value of $T$ depends on $i$. For simplicity, we slightly abuse notation and use the same $T$ for all blocks.}
For a block $U_i$, we start with character representations of the words, where
$\bar{x}_t^i = \textsc{CNN}_{\text{char}}(x_t^i).$
Then the character-based word representations are processed by the local Bi-LSTM, where
    $\{h_1^i,\ldots, h_T^i\} = \textsc{Bi-LSTM}_{\text{Local}}(\{\bar{x}_1^i,\ldots, \bar{x}_T^i\})$.

Before generating the globally-contextualized representations, the encoder  generates representations of all blocks by pooling the locally-contextualized word representations generated from the local encoder. The block representations are obtained using both max and average-pooling on the word representations, followed by a feed-forward transformation:
\begin{equation*}
    \begin{split}
    c_i = \textsc{FFNN}(&\textsc{Pool}_{\text{max}}(\{h_1^i,\ldots, h_T^i\});
                         \textsc{Pool}_{\text{avg}}(\{h_1^i,\ldots, h_T^i\})).
    \end{split}
\end{equation*}

The resulting local block representations $c_i$ are then processed by the global Bi-LSTM to generate document-contextualized block representations $p_i$:
$\{p_1,\ldots, p_K\} = \textsc{Bi-LSTM}_{\text{Global}}(\{c_1,\ldots, c_K\})$.

In downstream tasks, this document representation is used by inputting the text block contextual vector representations $\{p_i\}$ to light-weight task-specific networks.

\paragraph{\lrlstm}
As previously mentioned, when using left-to-right language models to pre-train
 representations, the representation
of the $i$-th word can only depend on the representations of the previous
words  $(i-1)$  words. Similarly, the representation
of the $j$-th text block can only depend on the representations of the previous text blocks. To design a document-level representation similar to the one defined by the \bilstm but that can still be trained using a uni-directional language model decomposition,  we  replace all of the bi-directional LSTMs in \bilstm with either
left-to-right or right-to-left LSTMs. The left-to-right version
of the resulting hierarchical encoder is presented in Figure~\ref{fig:left-to-right-hlstm}.

Given a document  $D = \{U_1, U_2, \ldots U_K\}$, 
let the encoding results of the left-to-right HLSTM  be denoted as
$\{\plr_1,\ldots, \plr_K\} = \textsc{L-to-R-HLSTM}(D; \theta_l).$
We use another right-to-left hierarchical encoder to produce representations:
$\{\prl_1,\ldots, \prl_K\} = \textsc{R-to-L-HLSTM}(D; \theta_r).$
We  then concatenate the final representations of the two encoders:
$\hat{p}_i = \begin{bmatrix} \plr_{i} &\prl_{i} \end{bmatrix}$.

We term the resulting document-level encoder~\lrlstm. Note that the \lrlstm combines the left and right information only at the top level, and there is no interaction between the two uni-directional contextualizers. While at the lowest layer, the \bilstm similarly concatenates left and right information, the second local {\sc bi}-LSTM layer can fuse the two directions. Similarly, the left and right contexts at the text segment level are fused at the second global {\sc bi}-LSTM layer.

In principle, both \bilstm and \lrlstm can capture long distance dependencies via their hierarchical structure, but it is unclear whether such dependencies can be effectively learned from limited amounts of supervised data labeled for down-stream tasks of interest. In the next section, we discuss how such document encoders can be trained to  provide rich text segment and token representations using large unlabeled document collections.

\section{Pre-training Hierarchical Document-Level Representations}
\label{sec:pre-training}

Here we propose language model-based methods for pre-training hierarchical document-level representations.
In order to train the hierarchical representations with
language models, our main idea is conceptually simple:
we combine the sentence representations
contextualized with respect to the whole document {\em
and} the token representations
with respect to the sentence to perform missing word predictions. Intuitively, from the loss of word prediction, both the contextual sentence
and token representations will be updated.

Note that there
are interactions between the contextual sentence and token
representations, as part
of the contextual sentence representations are constructed from token representations. Therefore, the
language model pretraining algorithms need to be designed carefully.
 In Section~\ref{sec:pre-training-uni}, we present ways to pre-train uni-directional hierchical document-level representations. In Section~\ref{sec:pre-training-bi}, we describe a language prediction model that can pre-train bi-directional hierarchical representations.

\subsection{Uni-directional Representations}
\label{sec:pre-training-uni}

Standard language model pre-training (\lm) can be used to pre-train the uni-directional token-level representations in local text block context. Starting from  Equation~\ref{eq:lr_lm_obj}, and using a standard LSTM language model,
we can define the predictive probability of the next word given prior text block-internal context as: $P^\text{local}_{\text{\llm}}(x_t^i|x_{t-1}^i, \ldots x_{1}^i) = P(x_t^i|\hlr_{t-1}^i)$, and train the representations using this language model. As in \newcite{peters-etal:2018:_deep}, this only pre-trains token representations in local context and does not provide a way to derive contextualized single vector text block representations.

Here we propose to use hierarchical language models to pre-train uni-directional hierarchical document-level representations.
We refer to this algorithm as \glm.
When pre-training with hierarchical language models, we treat the uni-directional document-level representation in Figure \ref{fig:left-to-right-hlstm} as the representation used by a hierarchical language model and use entire unlabeled documents to pre-train it.

The hierarchical language model predicts the next word in a document using the contextual encoder representations. The left-to-right hierarchical model
can use the contextual information from previous tokens in the same document
to predict the next word. The previous tokens with respect to a current token can be categorized into two sets: the preceding tokens in the same text block, and 
 all tokens in the {\em preceding} text blocks.
More specifically, the predictive probability for the next token is defined as:
    $ P^\text{global}_{\text{\llm}}(x_t^i|x_{t-1}^i, \ldots x_{1}^i, U_{i-1}, \ldots, U_1) = 
    P(x_t^i|\hlr_{t-1}^i, \plr_{i-1}).$
    
Therefore, the objective function for \glm is as follows:
\begin{equation*}
\begin{split}
     \min_{\theta_l, \theta_r} \sum_{t,i} & \log P(x_t^i|\hlr_{t-1}^i, \plr_{i-1}, \theta_l) + 
                               \log P(x_t^i|\hlr_{t+1}^i, \prl_{i+1}, \theta_r),
\end{split}
\end{equation*}
\vspace{-.05in}
where we use a feed-forward network to combine the information from $(\hlr_{t-1}^i, \plr_{i-1})$ and
$\hrl_{t+1}^i, \prl_{i+1}$, respectively. These feed-forward networks are not used in down-stream tasks.

\subsection{Bi-directional Representations}
\label{sec:pre-training-bi}

We have shown a method to pre-train uni-directional encoders with hierarchical language models. However, it is still not
clear how we can pre-train representations that do not satisfy the uni-directionality constraint from unlabeled text.
Here we adapt  masked language models  (\mlm)~\cite{devlin2018bert} to the task of pre-training hierarchical document representations. Masked language models alleviate the architectural constraints imposed by standard language model pre-training. They can be seen as a special form of denoising auto-encoders \cite{vincent-etal:2008:_extrac}, where the decoder network has minimal parameters. Unlike BERT \cite{devlin2018bert}, the  only type of noise we use is word masking, and the decoder only makes predictions for positions with noise (from the terminology in  \citet{vincent-etal:2008:_extrac}, full emphasis is on reconstructing the corrupted dimensions). 

\begin{wrapfigure}{rt}{0.5\textwidth}
\centering
  \includegraphics[width=6.5cm]{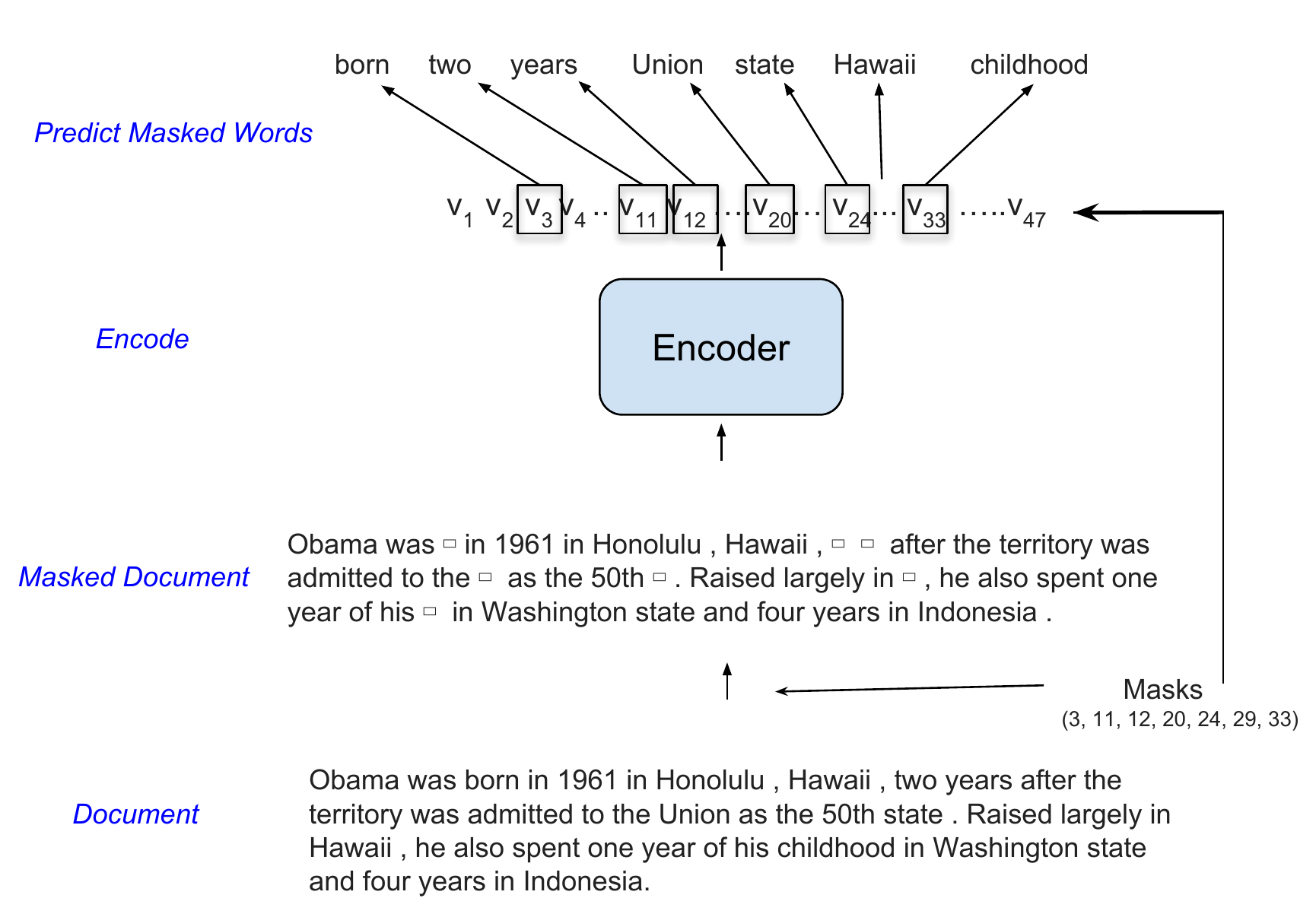}
\caption{The general procedure of using masked language model to train encoders. Note that masks are
randomly generated. See text for more details. }
\label{fig:mask_lm_procedure}
\end{wrapfigure}

Figure~\ref{fig:mask_lm_procedure} demonstrates the method of training representations with \mlm. 
For a given document, we first generate a set of random indices and mask out (hide) the words at the corresponding positions in the document. After a document has been masked, we provide it as input to a document-level encoder to produce contextual word and text block representations. These contextual representations are used to predict the words at the masked positions. 

More formally,  let
$Z$ denote  a set of indices for the masked words. 
Let $\zeta(D;Z) =\{\zeta(x_1^1), \zeta(x_2^1), \ldots, \zeta(x_t^i), \ldots \zeta(x_T^K)\} $ indicate the masked word sequence, where:
\begin{equation*}
  \zeta(x^i_t) =
\begin{cases}
  \rectangle, \forall (i, t) \in Z , \\
  x^i_t, \text{Otherwise.}
\end{cases}
\end{equation*}
Note that $\rectangle$ is a special symbol to indicate that the word is masked out (hidden).
Given that $Z$ is generated randomly,  masked language models essentially optimize
the following objective function
where $\theta$ represent the parameters of the representation:\footnote{Note that the model does not learn a probability distribution over possible texts and is thus not formally a language model. Nevertheless, for ease of exposition and in line with prior work on learning representations from unlabeled text \cite{Collobert:2008:UAN:1390156.1390177} we refer to this model as a language model.}
  $\max_{\theta}   \left( \E_{Z}  \left[ \sum_{(i,t) \in Z} \log P(x^i_t|\zeta(D;Z), \theta)\right]\right).$
  
Unlike the auto-encoder method of \citet{hill16}, in \mlm we assume the probabilities of all masked words in a sequence can be predicted independently and do not employ a separate auto-regressive decoder network.

There are two ways to apply \mlm to pre-train parts of the \bilstm.
Analogously to the local setting of \lm, we could pre-train with \mlm using local context.
We define the probabilities of masked words to depend only on the locally contextualized token representations, and apply 
\mlm to only pre-train the local bi-LSTM by using:
 $ P^{\text{local}}_{\text{\mlm}}(x^i_t|\zeta(D;Z); \theta) = P(x^i_t|h_t^i;\theta),$
where the $h_t^i$ is the word-level contextual representation for masked word $x_t^i$.
Note that if $(i,t) \in Z$, $\zeta(x_t^i) = \rectangle$. Therefore, $h_t^i$, which is generated by the
bi-directional local LSTM encoder, needs to carry the contextual information
in order to recover the masked word $x_t^i$. 

Similarly, in the \gmlm setting, the predictive probabilities are defined as:
 $ P^{\text{global}}_{\text{\mlm}}(x^i_t|\zeta(D;Z);\theta) = P(x^i_t|h_t^i, p_i;\theta),$
where we use another feed-forward network to combine
the information from $h_t^i$ and $p_i$.

%%% Local Variables:
%%% mode: latex
%%% TeX-master: "../meta"
%%% End:

\section{Down-stream Tasks}

We perform experiments on three document-level downstream tasks, which require predictions at the level of text blocks in full documents. The predictions are made using light-weight task specific networks which take as input single-vector document-contextualized text block representations $p_i$, generated by the hierarchical document encoders described in Section 3. We briefly define the downstream tasks and the task-specific architectures used. We provide additional details in the Experiments section.

\noindent{\textbf{Document Segmentation}}
In the document segmentation task, we take as input a document represented as a sequence of sentences. The goal is to predict, for each sentence, whether it is the last sentence of a group of sentences on the same topic, thereby segmenting the input document in topical  segments. We use a feed-forward network with one hidden layer with RELU activations on top of sentence representations $p_i$ to define scores of the task specific labels (segment boundary or not).

\noindent{\textbf{Answer Passage Retrieval}}
Given a document and a question, the task of answer passage retrieval is to select passages from the document that answer the question. While most QA systems focus on predicting short answer spans given a relevant answer passage,  answer passage retrieval is a prerequisite task for open-domain question answering.  While there are a multitude of architectures possible for this task, we propose a modular approach
that enables extremely fast answer passage retrieval during serving time. Our model works as follows.
The given document is encoded and its contextual passage representations $p_i$ are generated. The questions are treated as documents consisting of a single sentence and are similarly encoded in representations $q$. Inspired by \newcite{peters-etal:2018:_deep}, instead of using the representations $p_i$ directly, we use a linear combination of the layers of $p_i$ with positive weights summing to 1: $p_i' = \sum_l s_l p_{i,l}$, where $s_l$ is a task-specific learned scalar weight for the $l$-th layer, and $p_{i,l}$ represents
the $l$-th layer of the representations of the $i$-th block. To score a passage-question pair, we use the following architecture using a learned task-specific network
combining the following features: $s_0(p_i,q) = \langle p_i', q'\rangle$ is a dot product of the passage and question representations, $s_1,\ldots,s_5$ are five light-weight features such as paragraph position and tf-idf borrowed from \citet{clark-gardner:2018:_simpl}.  These features are sent to a feed-forward network with one hidden layer with RELU activations to generate the final score. 

\noindent{\textbf{Extractive Summarization}}
Given a document, the task of extractive summarization is to select representative sentences to form a summary for the whole document~\cite{Kupiec:1995:TDS:215206.215333}. Our summarization model is quite similar to our answer passage selection model with two key
differences. First, 
we add another randomly initialized LSTM on top of all sentence
representations to form refined sentence representations. We found that the extra layer generally improves results. Second, we perform max-pooling over the contextual representations of all sentences $p_i'$ to form a fixed-length vector representation $d$ for the entire document. The representation $d$ is then treated like the "question" representation in the passage selection task. Third, instead of using a dot-product as the scoring
function, for each sentence, we concatenate $d$ and $p_i'$ to form a combined vector which is sent to a feed-forward network with one hidden layer with RELU activations to generate the final score.  As for the Answer Passage Retrieval task, we add several task-specific features such as the position of the sentence in the document and the number of words in the sentence.

\section{Experiments}
The main goal of our experiments is to examine the value of
unsupervised pre-training of hierarchical document-level representations, focusing on the novel aspects of our methods. We first present the experimental settings
and pre-training details in Section~\ref{sec:exp-settings}. We then present and analyze the experimental results on the three tasks.

Among the pre-training methods,
\glm and \gmlm are proposed in this paper. The local version of \lm is
similar to ELMo~\cite{peters2017semi} and
 \mlm is similar to the method in~\citet{devlin2018bert}, but
the effectiveness of such methods when used to initialize document-level hierarchical representations has not been studied before. 

\subsection{Settings}
\label{sec:exp-settings}
We use standard LSTM cells for our sequence encoders. For word embeddings, we use character-based convolutions similarly
to~\cite{peters-etal:2018:_deep}, and project the embeddings to 512 dimensions. Our local encoder is a 2-layer LSTM with 1024 hidden units in each direction.
Our global encoder is another 2-layer LSTM with 1024 hidden
units.

Since \lm requires the representation to satisfy the uni-directionality constraint, while
there are no directionality constraints on the representations when pre-training with \mlm, in the experiments we always pair each encoder with its natural pre-training method: we pre-train \bilstm with \mlm, and  \lrlstm with \lm. 
For each of the two encoders, we compare no pre-training on unlabeled data, versus pre-training at the local sentence level only, versus pre-training at the global document level.

If we initialize the \bilstm representation with locally pre-trained \mlm, this means that
we initialize the local Bi-LSTM and learn the global Bi-LSTM only based on the labeled down-stream task data. If we initialize
both the local and global Bi-LSTMs of \bilstm with \gmlm pre-trained models, only the task-specific feed-forward network is randomly initialized, and all parameters are fine-tuned with the labeled set.
When comparing hierarchical (global) and non-hierarchical (local) pretraining
methods, we use the {\em same} document-level representation architecture for the downstream tasks, pre-train on the {same} amount of unlabeled text, and then
fine-tune 
with the same task-specific network architectures.

In all tasks, we also
include comparisons to skip-thought vectors~\cite{kiros-etal:2015:_skip}, where the model is learned to construct sentence embeddings.\footnote{We use the bi-directional model for the skip-thought vectors.} Additionally, we compare to \elmop,
 which forms a representation for a text block by pooling the token-level representations generated by the model pretrained in~\cite{peters-etal:2018:_deep}. 
\newcite{perone-silveira-paula:2018:_evaluat} have shown that \elmop produces strong sentence representations and outperforms many other sentence representation models including skip-thought vectors.\footnote{We use the "all layers" and "original" settings
according to \cite{perone-silveira-paula:2018:_evaluat}.} At a high level, the contextualized token representations obtained by our local \lm implementation are similar to the ones obtained by ELMo. However, ELMo was trained on a larger dataset and uses higher-dimensional hidden vector representations. Following prior work, skip-thought and ELMo vectors are used as fixed features and are not fine-tuned.

To pre-train the hierarchical representations, we use
documents in Wikipedia and filter out
ones with fewer than three sections or less than 800 tokens. We sample from the filtered document set and form a collection containing approximately 320k documents with about 0.9
billion tokens. We select the most frequent 250k words as the
vocabulary for all models. We limit documents to at most 75 sentences and limit each sentence length to 75 tokens. Sentences are used as text blocks for pre-training and pre-trained models are applied to both sentence-block and paragraph-block documents in the down-stream tasks. To capture the full hierarchical information, we did not use truncated back-propagation-through-time (tBPTT)~\cite{rumelhart-hinton-williams:1985:_learn}, but used full back propagation on documents with up to 5k tokens. For \mlm,  we randomly mask 20\% of the words in the documents for training. All of the down-stream task experiments
are performed using the same set of pre-trained models.

\subsection{Document Segmentation}
\label{sec:exp-downstream-seg}

We create labeled data for the segmentation
task by taking Wikipedia articles and forming the labels using the section information.
Given a sentence in a Wikipedia article, the label of the sentence is positive if it is the last sentence of a Wikipedia section.
To ensure fair comparisons, the section boundary information
is never used during pre-training. We select 5k documents each for
training, development, and test sets. \eat{Note that  positional information is important for the segmentation task, e.g. the last sentence of a document is always the last sentence of a section. Therefore, we remove the first three and the last three sentences in each document,  making the task more challenging.} In our dataset, only 5\%
of the sentences are positively labeled; we use  F1 score  as the evaluation metric due to this imbalanced class distribution.
All reported experimental results for the segmentation task are averages of five different training runs on the task-specific labeled data.

\paragraph{Results}
%%% Left 
The experimental results for the segmentation experiments are in
Table~\ref{tab:seg}. Our baseline system that does not use any pretraining of representations is similar to the one proposed in~\newcite{koshorek-etal:2018}
with the difference that our system uses character information to generate the word embeddings.
Note that pretaining results in substantial improvements over the baseline systems in all settings.
We next compare pre-training with hierarchical language models (the global setting) versus
pre-training with non-hierarchical language models (the local setting). 
Global pre-training of the hierarchical document-level representations  improves upon local pre-training of the sentence-contextualized token representations. In the case of the \lrlstm encoder paired with \glm, the improvement is over 6\% F1. For this task, we found that
\glm is much better than \gmlm. We hypothesize that by predicting words
in the next sentence, the \glm model is more sensitive to the topical changes between
sentences. The Table also shows results of experiments with a setting where we applied a Bi-LSTM on top of the \elmop and skip-thought vectors; as seen both methods under-perform  both our local and global pre-training approaches. In addition to not providing a method to pre-train the higher-level LSTMs, these methods do not fine-tune the local paragraph representations on downstream task data.

\begin{table*}[t]
\begin{subfigure}[b]{0.45\textwidth}
\hspace{-.2in}
\small
 \begin{tabular}{@{}lcc@{}}
     \toprule
                & \multicolumn{1}{c}{\mlm} & \multicolumn{1}{c}{\lm}            \\       
    %\midrule
    Pre-training & \bilstm               & \lrlstm              \\
     \midrule
     No         & 42.0 (0.0)       & 41.7 (0.0)        \\
     Local      & 50.6 (8.6)       & 48.4 (6.7)        \\
     Global     & {\bf 51.8} (9.8) & {\bf 54.9} (13.2) \\
     \midrule
     {\footnotesize LSTM+ELMo$_{\text{pool}}$} & \multicolumn{2}{c}{44.6} \\
     {\footnotesize LSTM+Skip-Thought} & \multicolumn{2}{c}{46.0} \\
     \bottomrule
   \end{tabular}
   \caption{Document Segmentation}
   \label{tab:seg}    
\end{subfigure}
\hspace{.5cm}
\begin{subfigure}[b]{0.45\textwidth}
 \begin{tabular}{@{}lcc@{}}
     \toprule    
     & \multicolumn{1}{c}{\mlm} & \multicolumn{1}{c}{\lm}            \\  
    Pre-training & \bilstm & \lrlstm \\ 
    \midrule
     No     & 77.24 (0.0)        &  77.20 (0.0)\\ 
     Local  & 79.17 (1.9)        &   78.36 (1.2)\\ 
     Global & {\bf 79.92} (2.7)  &  {\bf 79.57} (2.4)\\  
    \midrule
    {\footnotesize \cite{clark-gardner:2018:_simpl}}   & \multicolumn{2}{c}{73.31}           \\
    \bottomrule
    \end{tabular}
\caption{Answer Passage Retrieval}
\label{tab:triviaqa}
\end{subfigure}
\caption{
Downstream task performance for the segmentation and answer passage retrieval tasks. The improvements over corresponding baselines are indicated in brackets. {\bf (a)} F1 on document segmentation. {\bf (b)} Precision at one (P@1) for answer passage retrieval in TriviaQA-Wiki. We compare our
models to the
  answer passage retrieval module developed by \citet{clark-gardner:2018:_simpl}.}
\end{table*}

\subsection{Answer Passage Retrieval}
\label{sec:exp-downstream-triviaqa}

We apply the document-level encoders to the task of answer passage retrieval
in the next set of experiments. 
To evaluate the impact of pre-training a document-level encoder for this task, we use  the Wikipedia subset of the TriviaQA dataset~\cite{joshi-etal:2017:_triviaq}, a large scale
distantly supervised question answering dataset. 
While TriviaQA was originally designed for short answer detection, we use it to evaluate methods on the answer-passage selection task, as it requires retrieving relevant passages from one or more full documents as an important sub-task. \footnote{We did not study the TriviaQA Web subset as the first paragraph baseline has a strong accuracy of 83\% for it, reducing the importance of strong passage selection systems in this dataset.   }

We use the paragraph breaker implementation used in~\newcite{clark-gardner:2018:_simpl}. As in other work on TriviaQA, we set 400 to be the maximum number of tokens in a paragraph during training. We also use 400 tokens in testing. \eat{and 800 tokens for the results in Table \ref{tab:top-1-short-answers}. }  We report the top-1 passage selection accuracy on the development set as our main metric. As the TriviaQA test data labels are not publicly available, we cannot obtain passage accuracy test metrics and thus do not report these.

\begin{table*}[t]
\begin{subfigure}[t]{0.45\textwidth}
 \begin{tabular}{@{}lcc@{}}
     \toprule
                & \multicolumn{1}{c}{\mlm} & \multicolumn{1}{c}{\lm}            \\       
    %\midrule
    Pre-training & \bilstm               & \lrlstm              \\
     \midrule
     No         & 37.2  & 37.1    \\
     Local      & 37.4  & 37.3    \\
     Global     & {\bf 37.6} & {\bf 37.4}  \\       
     \bottomrule
   \end{tabular}
   \caption{Evaluating the impact of different pretraining algorithms. Rouge-L only is shown for brevity here.
   }
   \label{tab:summarization-pretrained}    
\end{subfigure}
\hspace{1cm}
\begin{subfigure}[t]{0.45\textwidth}
   \scalebox{.95}{
   \begin{tabular}{@{}lc@{}}
     \toprule    
    Pre-trained Vectors & P. Retrieval (P@1) \\ 
    \midrule
     \gmlm    & {\bf 66.2}\\ 
     \glm     &63.3\\ 
     \loclm$_{\text{pool}}$ & 53.8 \\
     \elmop   &{56.8}\\  
     Skip-Thought & 37.2 \\
    \bottomrule
    \end{tabular}
    }
\caption{To inspect the quality of the pretrained representations,
we show a zero-shot setting for the answer passage retrieval task.}
\label{tab:triviaqa-fixed}
\end{subfigure}
\caption{Analysis of the impact of different pretraining methods on document summarization and zero-shot answer passage retrieval.}
\end{table*}

\begin{table}[t]
  \ra{1.3}
   \centering
   \scalebox{.85}{
   \begin{tabular}{@{}lcccc@{}}
     \toprule    
    Models & \multicolumn{1}{c}{R-1} & \multicolumn{1}{c}{R-2}
     & \multicolumn{1}{c}{R-L} \\  
    \midrule
    LEAD \cite{see2017get} & 39.6	& 17.7 &	36.2 & \\
    NeuralSum \cite{cheng-lapata:2016:P16-1}    & 35.5        &  14.7 & 32.2 \\
    SummaRuNNer \cite{Nallapati17}  & 39.6 & 16.2 &35.3 \\
    REFRESH \cite{narayan2018ranking} & 40.0 & 18.2 & 36.6 \\
    Bottom-up Summarization \cite{gehrmann2018bottom} [Extractive] & 40.7 &18.0 &37.0\\
    \gmlm ({\bf Ours}) & {41.2} & {\bf 19.1} & {37.6} \\
    \midrule
    LSTM$+$Elmo$_\text{pool}$ & 41.0	&18.9 &	37.4 \\
    LSTM$+$Skip-Thought & 40.8 &	18.7&37.2\\
    \midrule
    NeuSum \cite{zhou2018neural} & {\bf 41.6} & 19.0 & {\bf 38.0}\\
    \bottomrule
    \end{tabular}
    }
  \caption{Comparative evaluation of our summarization models with respect to other extractive summarization systems. ROUGE-1 (R-1), ROUGE-2 (R-2), and ROUGE-L
(R-L) F1 scores are reported. The first part of the table indicates system results with
non-autoregressive sentence selection models, including prior work and our best proposed approach. The second
part of the table shows the performance of (non-autoregressive) models using pre-trained sentence embeddings input to a document-level Bi-LSTM. NeuSum~\cite{zhou2018neural} is included in the third part as it is specifically trained to score combinations of sentences with auto-regressive selection models.}
  \label{tab:summarization-compared}
\end{table}

The passage retrieval results are in Table~\ref{tab:triviaqa}.  Note that pre-training leads to substantial improvements in all settings. For both \mlm and \lm, global pre-training of the full hierarchical representations improves upon local pre-training of the lower-level contextual token representations. 
Our models improve upon the passage selection model proposed by \citet{clark-gardner:2018:_simpl} by more than 6\% absolute. The latter work focuses on evaluation of their end-to-end short answer selection system and does not report passage retrieval performance for TriviaQA-Wiki. We measured the performance using the authors' codebase\footnote{\url{https://github.com/allenai/document-qa}}. 

\noindent \textbf{Analysis}
Our answer passage retrieval model uses a dot-product
between the question and passage representations to score passages. If we restrict our models to not use additional features apart from this dot product of the question and passage representations, we can derive task-specific models in a "zero-shot" setting, where we do not use labeled data from TriviaQA to fine-tune pre-trained representations. We use this zero-shot setting to further
investigate whether the pre-trained representations have learned to assess the semantic similarity between questions and passages based on unlabeled text. 
The results are presented in Table~\ref{tab:triviaqa-fixed}.
In this setting, \gmlm achieves 66.2, outperforming \glm, \elmop and skip-thought vectors. The max-pooled version of our local \lm method under-performs \elmop, which is explained by the difference in size of the pre-trained representations.

\subsection{Extractive Document Summarization}
\label{sec:exp-downstream-sum}

We use the CNN/DailyMail dataset~\cite{hermann-etal:2015:_teach} to evaluate our summarization model. We follow the standard data pre-processing and evaluation settings for extractive summarization from \cite{see2017get}, and use ROUGE~\cite{lin2004rouge} as the evaluation metric. In the training data, there are only gold summaries without sentence labels. To generate labels for sentence selection, we follow~\citet{Nallapati17} to create extractive oracle summaries by starting with an empty summary and greedily adding one sentence at a time (up to a total of three sentences) to optimize the ROUGE score between the selected sentences
and the gold summarizes. Our models are trained to choose
these selected sentences. We add three loss functions during training for selecting the first, second, and third sentence, respectively. The ROUGE score gain attributed to each sentence is used to weight the loss functions. During testing, we simply select the top three sentences according to our models.

The results of our summarization system are presented in Table~\ref{tab:summarization-compared}. Compared to  other
extractive summarization systems, our model improves upon the prior
work of~\citet{narayan2018ranking} by 0.9 in ROUGE-1 and 0.7 in ROUGE-L, without using the sophisticated reinforcement learning techniques employed by that work. Our model is competitive to NeuSum~\cite{zhou2018neural} for ROUGE-2, but has worse ROUGE-1 and ROUGE-L. We hypothesize that this
is due to the fact that NeuSum~\cite{zhou2018neural} employed
an auto-regressive sentence selection model. Recently, abstractive summarization systems were able to outperform extractive ones \cite{paulus2017deep,gehrmann2018bottom}.
We expect that the novel aspects of our work can also bring improvements in that non-extractive setting and are complementary to the use of auto-regressive sentence selectors.
We also compared our models to models using \elmop
and skip-thought vectors as fixed features input to a document-level Bi-LSTM. As seen these methods are also competitive, but are not as strong as our global hierarchical pre-training approach.

The impact of  pretraining is analyzed in Table~\ref{tab:summarization-pretrained}.
For brevity, we only include ROUGE-L scores, but ROUGE-1 and ROUGE-2 show similar trends. As for answer passage retrieval, \gmlm
is the best overall pretraining algorithm. 

%%% Local Variables:
%%% mode: pdflatex
%%% TeX-master: "../meta"
%%% End:

\section{Related Work}

Prior work in {\em supervised} learning of neural representations for full documents has shown the effectiveness of hierarchical neural structures, which contain representations of both the sentences and their component words in a two-level hierarchy. In
the lower hierarchy level, the contextual word representation
is formed with respect to the sentence. In the higher level, the contextual sentence representation is formed with respect to the document.
Such structures enable the models to integrate long-distance context from the documents and have been used for labeling sentence sentiment \cite{ruder2016hierarchical}, document summarization \cite{cheng-lapata:2016:P16-1}, text segmentation~\cite{koshorek-etal:2018}, and text classification~\cite{Yang2016HierarchicalAN}, \emph{inter alia}. These hierarchical neural representations have been largely learned based on task-specific labeled data, posing a challenge for applications with a limited number of annotated examples.  

Unsupervised pretraining for hierarchical document representations has received
relatively little attention despite the recent success on language model pre-training~\cite{peters-etal:2018:_deep, radford-etal:2018}. 
Prior work has shown how to use unlabeled text to pre-train representations of individual words, e.g. \cite{word2vecNIPS2013_5021}, or flat (relatively short) sequences of words, where each word representation is contextualized with respect to the sequence \cite{peters2017semi,peters-etal:2018:_deep,salant2018contextualized,radford-etal:2018}. 
%Such contextual representations have brought much larger improvements in downstream tasks, by helping learn to integrate contextual information within a sequence. 
Similarly, fixed-length vector representations of full sentences/paragraphs have also been pre-trained from unlabeled text \cite{le-mikolov:2014:_distr, kiros-etal:2015:_skip,dai2015semi,logeswaran2018efficient}, where the sentence/paragraph representations are generated based on the content of the sentence/paragraph alone, ignoring other
sentences in the document.\footnote{Note that most approaches (e.g. \cite{kiros-etal:2015:_skip}), pretrain the representations by asking a model to predict the words in other sentences during training, but the encoders look at sentences in isolation to generate representations at inference time.}

\citet{li-luong-jurafsky:2015:ACL-IJCNLP} showed how to train hierarchical document representations from unlabeled text using document auto-encoders but did not use such representations in extrinsic downstream document-level tasks. To the best of our knowledge, no prior work has transferred hierarchical unsupervised document representations to downstream tasks, and evaluated the impact of pretraining local and global document-level contextualizers. In addition, bidirectional hierarchical representations have not been pretrained without a sophisticated decoder in an auto-encoder framework before.

\section{Conclusion}

In this paper, we proposed methods for pre-training  hierarchical document representations, including contextual token and sentence/paragraph representations, integrating context from full documents. We demonstrated the impact of pre-training such representations on three document-level downstream tasks: text segmentation, passage retrieval for document-level question answering, and extractive summarization.

\bibliographystyle{iclr2019_conference}
\bibliography{lumiere}
\end{document}